\pdfoutput=1

\documentclass[11pt]{article}

\usepackage{acl}

\usepackage{times}
\usepackage{latexsym}

\usepackage[T1]{fontenc}

\usepackage[utf8]{inputenc}

\usepackage{microtype}

\usepackage{colortbl}
\usepackage{graphicx,multirow}

\newcommand{\BPEs}{\texttt{BPEs}}
\newcommand{\Morfessor}{\texttt{Morfessor}}
\newcommand{\semiCRFs}{\texttt{semiCRFs}}
\newcommand{\CRFs}{\texttt{CRFs}}
\newcommand{\flatcat}{\texttt{FC}}
\newcommand{\LMVR}{\texttt{LMVR}}
\newcommand{\seqtoseq}{\texttt{s2s}}
\newcommand{\seqtoseqmtrand}{\texttt{s2s+multi}}

\newcommand{\ptrseg}{\texttt{PtrSeg}}

\title{BPE vs. Morphological Segmentation: A Case Study on Machine Translation of Four Polysynthetic Languages}

\author{ \textbf{Manuel Mager${ }^{\diamondsuit}$ \quad
Arturo Oncevay${ }^{\heartsuit}$ \quad
Elisabeth Mager${ }^{\sharp}$ }\\
\textbf{Katharina Kann${ }^{\spadesuit}$
Ngoc Thang Vu${ }^{\diamondsuit}$  } \\
${ }^{\diamondsuit}$University of Stuttgart \quad
${ }^{\sharp}$Universidad Nacional Autónoma de México \\
${ }^{\spadesuit}$University of Colorado Boulder \quad
${ }^{\heartsuit}$University of Edinburgh }

\begin{document}
\maketitle
\begin{abstract}
Morphologically-rich polysynthetic languages present a challenge for NLP systems due to data sparsity, and a common strategy to handle this issue is to apply subword segmentation. We investigate a wide variety of supervised and unsupervised morphological segmentation methods for four polysynthetic languages: Nahuatl, Raramuri, Shipibo-Konibo, and Wixarika. Then, we compare the morphologically inspired segmentation methods against Byte-Pair Encodings (BPEs) as inputs for machine translation (MT) when translating to and from Spanish. We show that for all language pairs except for Nahuatl, an unsupervised morphological segmentation algorithm outperforms BPEs consistently and that, although supervised methods achieve better segmentation scores, they under-perform in MT challenges. Finally, we contribute two new morphological segmentation datasets for Raramuri and Shipibo-Konibo, and a parallel corpus for Raramuri--Spanish.
\end{abstract}

\section{Introduction}

Polysynthetic languages are known because of their rich morphology, that encodes most parts of the semantics into verbs, leading to a high morpheme-per-word rate. The resulting combinations of morphemes and roots  
result in extreme type sparsity. Thus, polysynthetic languages represent a challenging environment for NLP methods \cite{klavans-2018-computational}.  
Subword segmentation has been a common method to reduce sparsity \cite{vania2017characters}. Moreover, as these languages are mostly extremely low-resource (ELR), the challenge is even harder. Some of the reasons behind this is that most of them are endangered and spoken by minority groups \cite{mager-etal-2018-challenges,littell-etal-2018-indigenous}.   

But what impact does morphological segmentation have on downstream tasks like machine translation (MT), when translating from or into fusional languages? Linguistically inspired segmentation was considered to be the best option to handle rich morphology \cite{koehn2005europarl,virpioja2007morphology} until the appearance of Byte-Pair Encodings \citep[BPEs;][]{sennrich2016neural} and has been adopted as the default segmentation technique. BPEs earned this status for its good results, unsupervised training and language independence.  \newcite{saleva-lignos-2021-effectiveness} show that there is no significant gain when using an unsupervised morphological segmentation for the input over BPEs when evaluating those methods in moderate LR scenarios for Nepali--English and Kazakh--English, contradicting initial findings of \newcite{ataman-federico-2018-evaluation}. However, how would BPEs perform for polysynthetic languages in ELR scenarios? \newcite{schwartz2020neural} compare BPE, with Morfessor \cite{smit-etal-2014-morfessor} and Rule-Based morphological analyzers for medium resourced Inuktitut--English, and for the ELR Yupik--English and Guarani--Spanish. Their results show that BPEs outperform Morfessor and the morphological analyzer in all MT cases (but with better Language Modeling capabilities of morphological models over BPEs).  
However, most of these studies only rely on the usage of a limited set of segmentation methods and do not consider the quality of the used morphological segmentation methods.

This study aims to answer the following research questions: i) is morphological segmentation beneficial for MT where one language is polysynthetic and ELR?; and ii) is higher morphological segmentation quality correlated with higher MT scores?

To answer these questions, we perform segmentation experiments on four polysynthetic languages:\footnote{We choose the languages for this study based on the availability of a morphological segmentation dataset.} Nahuatl (\texttt{nah}), Raramuri (\texttt{tar}), Shipibo-Konibo (\texttt{shp}) and Wixarika (\texttt{hch}) and apply those segmentations to MT paired with Spanish (\texttt{spa}). First, we revisit a wide set of supervised and unsupervised methods and apply them to the input of MT transformer models. This study is the first to show that strong unsupervised morphological approaches outperform BPEs consistently on ELR polysynthetic languages, except for \texttt{nah}. These results are related to \newcite{ortega2020neural}, that found that a morphologically guided BPE can improve the MT performance for Guarani--Spanish. On the other hand, even when supervised morphological segmentation methods achieve better results for the segmentation task, when it comes to MT systems they under-perform all other approaches.  We hypothesize that this might be due to overfitting the clean and out-of-domain morphological training set. To make all these experiments possible we introduce additionally two new morphologically annotated datasets for \texttt{tar} and \texttt{shp}; and one parallel dataset for \texttt{spa}--\texttt{tar}\footnote{The datasets are available under \url{http://turing.iimas.unam.mx/wix/mexseg}}.

\begin{table}[t]
\centering
\setlength{\tabcolsep}{3pt}
\resizebox{0.95\linewidth}{!}{%
\begin{tabular}{l|cc|cc|cc}
         & \multicolumn{2}{c|}{train} & \multicolumn{2}{c|}{dev}   & \multicolumn{2}{c}{test}  \\
         & tar         & spa          & tar         & spa          & tar          & spa         \\ \hline
S        & \multicolumn{2}{c|}{13,102} & \multicolumn{2}{c|}{587}   & \multicolumn{2}{c}{1,030}  \\
N\textsubscript{spa}/N\textsubscript{tar} & \multicolumn{2}{c|}{1.692} & \multicolumn{2}{c|}{1.794} & \multicolumn{2}{c}{1.689} \\
N        & 73,022       & 93,410       & 3,183        & 4,133        & 5,847         & 7,547       \\
V        & 19,044       & 16,220       & 1,713        & 1,771        & 2,793         & 2,803       \\
V1       & 12,894       & 10,021       & 1,402        & 1,365        & 2,221         & 2,120       \\
V/N      & 0.261       & 0.174       & 0.538       & 0.429       & 0.478        & 0.371      \\
V1/N     & 0.177       & 0.107       & 0.440       & 0.330       & 0.380        & 0.281      \\
OOV      &             &             & 573         & 434         & 1,037         & 779  \\
\%OOV      &             &             & 0.334         & 0.245         & 0.371         & 0.277
\end{tabular}
}
\caption{Parallel corpus' description: S = number of sentences; N\textsubscript{spa}/N\textsubscript{tar} = ratio of tokens between Spanish and Rar\'amuri; N = number of tokens; V = vocabulary size; V1 = number of tokens occurring once (hapax); V/N = vocabulary growth rate; V1/N = hapax growth rate; OOV = out-of-vocabulary words w.r.t. train set.}
\label{tab:parallel-corpus}
\end{table}

\paragraph{Polysynthetic languages.}
\label{sec:poly}
A polysynthetic language is defined by the following linguistic features: 
the verb in a polysynthetic language must have an agreement with the subject, objects and indirect objects \cite{baker1996polysynthesis};
nouns can be incorporated into the complex verb morphology \cite{mithun1986nature}; and, therefore, polysynthetic languages have agreement morphemes, pronominal affixes and incorporated roots in the verb \cite{baker1996polysynthesis}, and also encode their relations and characterizations into that verb.

 \begin{table}[]
\centering
\setlength{\tabcolsep}{3.5pt}
\resizebox{0.95\linewidth}{!}{%
\begin{tabular}{l|ccc|ccc}
           & \multicolumn{3}{c|}{\texttt{shp}} & \multicolumn{3}{c}{\texttt{tar}} \\
           & train   & dev   & test  & train   & dev   & test  \\ \hline
Words      & 604     & 163   & 329   & 504     & 136   & 274   \\
SegWords   & 437     & 114   & 228   & 323     & 87    & 178   \\
Morphs     & 1215    & 321   & 642   & 1028    & 273   & 563   \\
UniMorphs  & 476     & 181   & 319   & 474     & 181   & 287   \\ \hline
Seg/W      & 0.72       & 0.69     & 0.69     & 0.64    & 0.64  & 0.65  \\
Morphs/W   & 2.01    & 1.97  & 1.95  & 2.04    & 2.01  & 2.06  \\
MaxMorphs  & 5       & 5     & 5     & 5       & 5     & 5     \\
OOV-M &         & 93    & 179   &         & 93    & 163  
\end{tabular}
}
\caption{Number of words, segmentable words (SegWords), total morphemes (Morphs), and unique morphemes (UniMorphs) in our new datasets. Seg/W: proportion of words consisting of more than one morpheme; Morphs/W: morphemes per word; MaxWords: maximum number of morphemes found in one word; OOV-M: morphemes in evaluation not seen in training.}
\label{tab:seg-data-tar-shp}
\end{table}

\section{Descriptions of Novel Datasets} 
\label{sec:datasets}

\subsection{Raramuri--Spanish Parallel Dataset}
\label{subsec:rarmuriparall}

For the dataset, we manually extract phrases that had a translation into Spanish from the \newcite{brambila1976diccionario} dictionary. Additionaly, given that the orthography  in this book is out of use, we normalized it to a modern version used in \cite{caballero2008choguita}. The book does not specify the dialect of the sentences. Table \ref{tab:parallel-corpus} shows the characteristics of the dataset, and the dataset splits.

\subsection{Morphological Segmentation Datasets}
\label{subsec:segmentationdata}

We also introduce two new morphologically annotated datasets. For Raramuri we manually extracted segmented morphemes from a specialized linguistics paper \cite{caballero2010scope} and thesis \cite{caballero2008choguita} that contain segmented and non-segmented words. Both sources annotate the Raramuri variant of the village of Choguita.

For Shipibo-Konibo, we adapted annotated sentences for lemmatization and part-of-speech tagging \cite{pereira-etal-2017-ship}, and from a treebank \cite{vasquez-etal-2018-toward}, which was segmented in morphemes due to a particular phenomenon for clitics in the dependencies annotation.

\section{Experimental Setup}

\subsection{Resources}
\label{subsec:useddata}
For the machine translation experiment we use the following parallel datasets: the \texttt{hch}--\texttt{spa} translation of the fairy tales of Hans Christian Andersen \cite{mager-wixarika}; the Shipibo-Konibo--Spanish translations from a bilingual dictionary and educational material \cite{galarreta-etal-2017-corpus}; and for \texttt{nah}--\texttt{spa}, the Axolotl dataset \cite{gutierrez-vasques-etal-2016-axolotl}. This dataset contains several variants of Nahuatl. On top of that we also use our collected \texttt{tar}--\texttt{spa} Parallel corpora (\S\ref{subsec:rarmuriparall}). The details of the data splitting are described in Table \ref{tab:data_splitting} in the appendix. For morphological segmentation we use the \texttt{nah} and \texttt{hch} annotated datasets from \citet{kann-etal-2018-fortification} and additionally we use the \texttt{shp} and \texttt{tar} datasets introduced in section \ref{subsec:segmentationdata}. We use the same splits as reported by the original sources.

\subsection{Metrics}

For machine translation we use the standard BLEU \cite{papineni2002bleu} and chrF \cite{popovic2015chrf} metrics from the SacreBLEU implementation \cite{post-2018-call}. To evaluate morphology, we compare all outputs against the gold annotated test sets calculating accuracy and the EMMA F1 metric \cite{spiegler-monson-2010-emma}.

\subsection{Subword Segmentation}
\label{subsec:subword}
\paragraph{BPEs} \citep[\BPEs; ][]{sennrich2016neural} is our reference system we use the sentence piece implementation  \cite{kudo-richardson-2018-sentencepiece} of BPEs. We tune the vocabulary size on a vanilla transformer small for each language, and take the best model evaluated on the development set.

\paragraph{Morfessor}\citep[\Morfessor; ][]{smit-etal-2014-morfessor}
As an unsupervised method we use Morfessor 2.0, that is a statistical model for the discovery of morphemes using  minimum description length optimization. 

\paragraph{FlatCat} \citep[\flatcat; ][]{gronroos2014morfessor}, is a variant of Morfessor. It consists of a category-base hidden Markov model and a flat lexicon structure for segmentation.

\paragraph{LMVR} \cite{ataman2017linguistically} modify the \texttt{FC} implementation by adding a lexicon size restriction and increase the tendency of the model to increase segmentation of commonly seen words.

\paragraph{CRFs}(\CRFs) As our first supervised model we use the  
conditional random fields \citep[CRFs; ][]{lafferty2001conditional} segmentation model of \newcite{ruokolainen-etal-2014-painless}.  
We also investigate the capabilities of \semiCRFs~ \cite{sarawagi2005semi} for this particular task. For this, we use the Chipmunk implementation \cite{cotterell-etal-2015-labeled}.

\paragraph{Seq2seq} We also use a vanilla RNN sequence-to-sequence model with attention. The first variant (\seqtoseq) employs a supervised neural model. Additionally, we use the most promising extension proposed by \newcite{kann-etal-2018-fortification} adding random generated strings in an auto-encoding fashion (\seqtoseqmtrand).

\paragraph{Pointer--Generator Networks} \citep[\ptrseg;][]{see-etal-2017-get} are commonly used in task where copying part of the input to the output is part of the task. This model has been used successfully for canonical segmentation \cite{mager-etal-2020-tackling}.

\begin{table}[b]
    \centering
        \setlength{\tabcolsep}{5.5pt}
    \begin{tabular}{c | r r r r}
        system&\texttt{hch}&\texttt{nah}&\texttt{tar}&\texttt{shp}\\\hline
\BPEs              & 53.17& 53.38& 62.54& 71.41\\\hline
\Morfessor        & 61.51& 60.48& 59.05& 59.45\\
\flatcat          & \underline{62.28}& 58.94& 64.65& \underline{67.95}\\
\LMVR             & 61.27& \underline{60.55}& \underline{65.46}& 67.58\\\hline
\semiCRFs          & 68.10& 81.92& 81.22&     -\\
\CRFs             & 82.43& \bf 87.83& 89.79&     -\\
\seqtoseq          & 82.42& 84.62& 88.47& 82.25\\
\seqtoseqmtrand& \bf 83.75& 84.90& 88.37& \bf 85.99\\
\ptrseg       & 65.60& 83.85& \bf 90.13& 78.22\\\hline\hline

    \end{tabular}
    \caption{Test results of surface segmentation  for \texttt{hah}, \texttt{nah} and \texttt{tar}, and canonical segmentation for \texttt{shp}. Values are F1 scores, bold numbers are the best systems overall, underscored are the best unsupervised systems.}
    \label{tab:morph}
\end{table}

\subsection{NMT System}
As our translation models, we use an encoder-decoder transformer model \citep{vaswani2017} with the hyperparameters proposed by \citet{guzman-etal-2019-flores} as a baseline for low-resource languages. We use the vanilla version of this transformer without any further back-translation or other enhancements, so that we can remove any additional variables from the experiment, and focus only on the input segmentation. We use a 5k\footnote{We searched for the best vocabulary size using 2k, 4k, 5k, 6k and 8k.} vocabulary size for all sides using BPE.  
We use fairseq \cite{ott-etal-2019-fairseq} for all translation experiments. The polysynthetic languages are segmented with the different investigated segmentation methods and Spanish always uses BPE in both translation directions.

\begin{figure*}
    \centering
    \includegraphics[width=.9\linewidth]{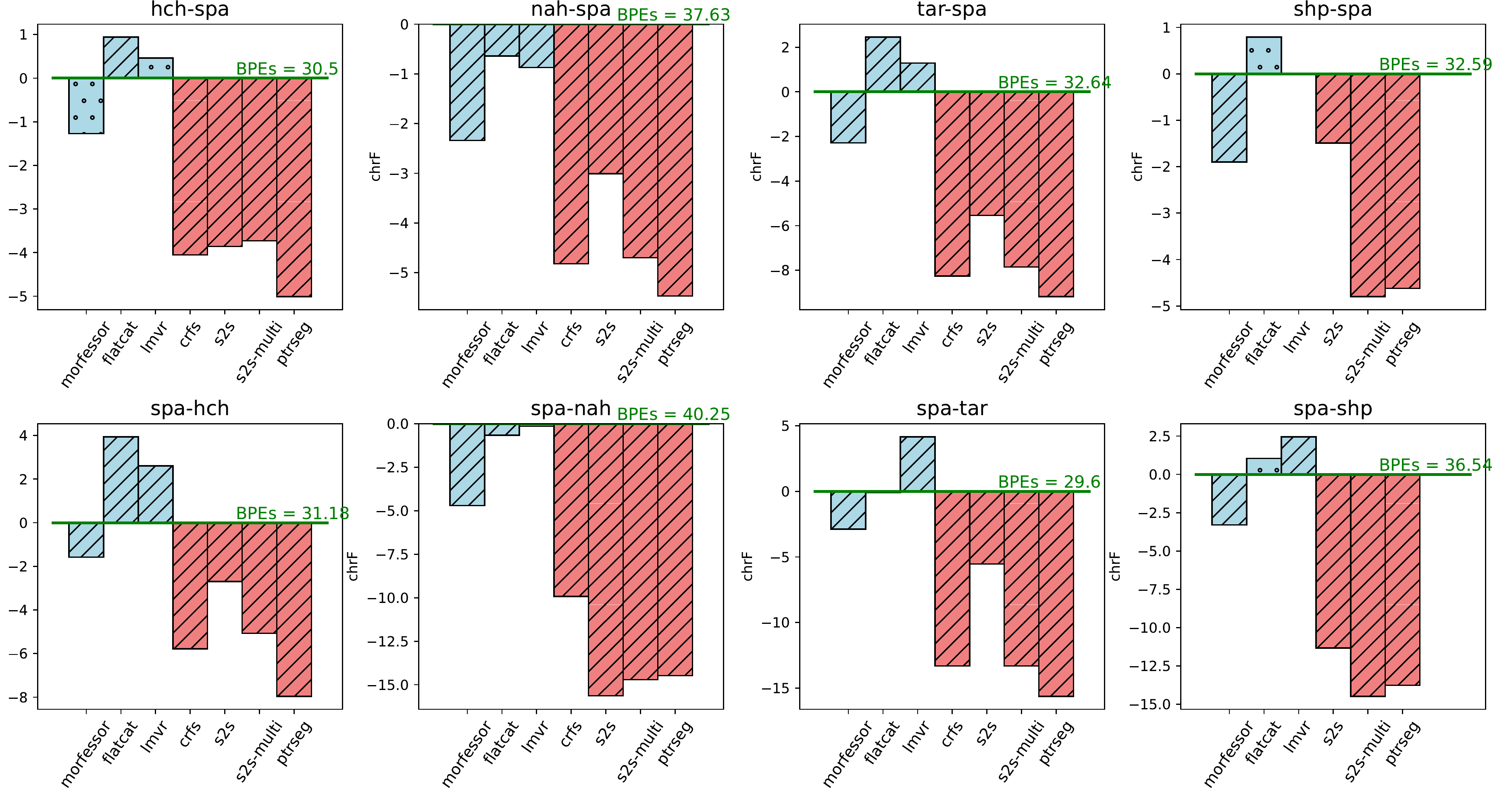}
    \caption{chrF score difference for all morphological segmentation when compared to \BPEs~ on the test sets for both translation directions. We run a paired approximation test with 10000 trials using the \BPEs~ system output as the baseline. Diagonals indicates a p-value $\leq$ 0.05, while stars indicates a p-value $>$ 0.05. Blue systems are unsupervised, while Red ones are supervised.}
    \label{fig:mt}
\end{figure*}

\section{Results}
\paragraph{Morphology} Table \ref{tab:morph} shows that \BPEs, a model that is not intended for morphological segmentation, perform worst on all languages as expected, with exception of \texttt{tar}. The unsupervised morphological segmentation models (\Morfessor, \flatcat~and \LMVR) are consistently the worst performing models among the morphologically inspired models. The best performing systems are supervised, with \seqtoseqmtrand~showing best results for \texttt{hch} ($83.7$ F$_1$) and \texttt{shp} ($85.99$ F$_1$). \CRFs~achieved the best result for \texttt{nah} with $87.8$ F$_1$ and \ptrseg~achieved the best scores for \texttt{tar} with $90.13$ F$_1$.

\subsection{Discussion}

\paragraph{MT} Figure \ref{fig:mt} shows the chrF score difference against the \BPEs~baseline in all directions\footnote{See Table \ref{tab:mt} for the specific scores, BLEU ones included.}. We first observe that the supervised segmentation approaches under-perform in contrast with the unsupervised ones in all the settings. 

Moreover, with the polysynthetic languages in the source side, \flatcat~has a significantly higher score for \texttt{hch-spa} and \texttt{tar-spa}, and a statistical tie in \texttt{shp-spa}; whereas \LMVR~obtains similar results to \BPEs~in \texttt{hch-spa} and \texttt{shp-spa}. In the other direction, with the polysynthetic languages as targets, \LMVR~is the method that significantly surpasses the baseline for more language pairs: \texttt{spa-hch}, \texttt{spa-tar} and \texttt{spa-shp}; whereas \flatcat~obtains the maximum score in \texttt{spa-hch} and statistical ties in \texttt{spa-tar} and \texttt{spa-shp}. We conclude that both methods are robust alternatives for translating from and to a polysynthetic language.

Despite the good results of \seqtoseq, \seqtoseqmtrand~or \ptrseg~in morphological segmentation, for MT they have the worst performance. We argue that these kind of methods innovate new subwords in their output, which can aid for morphological segmentation, but for MT only adds noise in the input for the model. 

Overall, we notice that in contrast to other languages \cite{saleva-lignos-2021-effectiveness}, segmentation methods matter for polysynthetic ones. Poor suited methods can strongly decrease the performance of down-stream tasks like MT. However, the question on which segmentation method is better for MT is still open. 

\subsection{Analysis}
\begin{figure*}[h]
    \centering
    \includegraphics[width=1\linewidth]{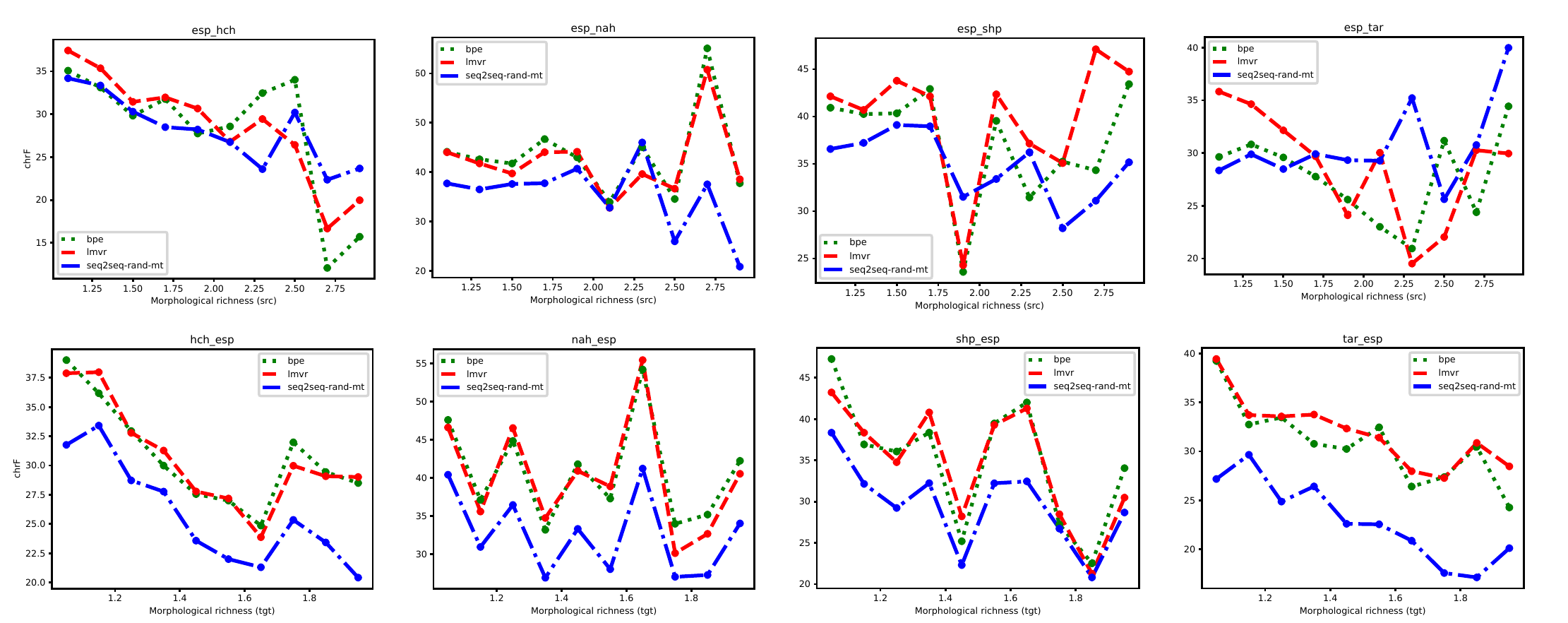}
    \caption{Relation between morphological richness of each polysyntetic language with relation to its chrF score, in each translation direction. The scores are analysed for \BPEs, \LMVR~ and \seqtoseqmtrand.}
    \label{fig:morphrich}
\end{figure*}

\begin{figure*}[h]
    \centering
    \includegraphics[width=1\linewidth]{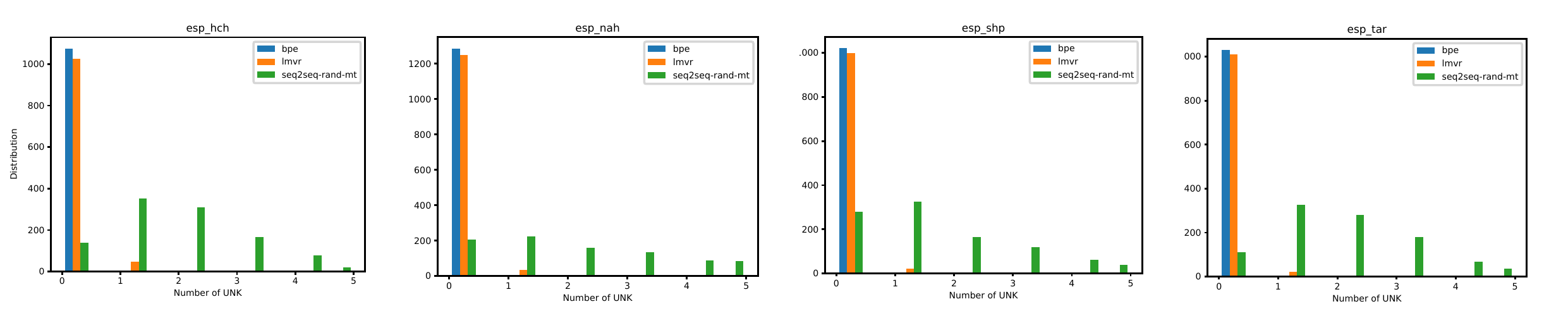}
    \caption{Number of out-of-vocabulary tokens (UNK) found for each polysynthetic language classified by system. The scores are analysed for \BPEs, \LMVR~ and \seqtoseqmtrand.}
    \label{fig:unk}
\end{figure*}
To better understand the current results, we explore the outputs of different systems. For simplicity, we choose the best performing segmentation system for each of the segmentation paradigms. For unsupervised morphological inspired segmentation, we use \LMVR, \seqtoseqmtrand~ for supervised morphological segmentation, and \BPEs~ for frequency-based segmentation.

First, we explore the impact of morphological richness on each of the systems. We use \Morfessor~ to infer the segmentation for each polysynthetic language data point and divide the number of found morphemes by the total number of tokens. Figure \ref{fig:morphrich} shows that there is no clear correlation between morphological richness and systems' performance for \texttt{nah} and for \texttt{shp}. However, for \texttt{hch} we observe that a richer morphology implies a loss in translation quality. The same correlation can be seen for the \texttt{tar-esp} direction. This correlation is stronger when the polysynthetic language is in the source and weaker when it is in the target. Overall, a similar behavior can be observed between \LMVR~ and \BPEs.

Second, we explore the impact of out-of-vocabulary (UNK) tokens that each segmentation model introduces because having a high number of UNK tokens can negatively influence the MT results. In figure \ref{fig:unk}, we show the number of UNK tokens that each segmentation has when used with the dictionary of an MT system. The supervised {\seqtoseqmtrand} has the highest amount of UNK symbols. We suggest that the reasons behind this phenomena could be the strong generative power of such systems and well-known artifacts that such models introduce (i.e., string repetitions). However, {\LMVR} has a slightly higher number of UNK tokens, leaving {\BPEs} the best vocabulary coverage. This can explain the surprisingly low performance of supervised models.

\section{Conclusion}

In this paper, we compared a wide set of morphological segmentation models with BPEs when applied to the input of Neural Machine Translation systems for extreme low-resource polysynthetic languages. We found that unsupervised morphological segmentation outperformed BPEs significantly on 5 out of 8 language pairs, setting a consistent overall performance. Surprisingly SOTA supervised morphological segmentation achieved the lowest performance of all systems. In future, we will explore Adaptor-Grammars \cite{johnson2006adaptor,narasimhan2015unsupervised,eskander-etal-2020-morphagram} for segmentation, and also the way to make unsupervised segmentation more robust and suitable for MT including the reduction of produced UNK symbols.

\section*{Ethical Considerations}

The datasets introduced in this paper for machine readable training and evaluations are extracted from previous specialized linguistic  work. We stick to the ethical standards giving credit to the original author in the spirit of \emph{fair scientific usage}. We further strongly encourage future work that use these resources to cite also the original sources of the data. Additionally we found another ethical risks of this work: for the down-stream task of MT, a translation system should not be deployed with low quality translations, as it can mislead the user, and have implicit biases. Finally, want to state that the authors of this paper have a long record of working with the studied indigenous languages. 
Some have conducted field studies with the communities in the past, and Manuel Mager is part of the Wixarika community. This allows the authors to have a better understanding of the concerns of the communities that speak the discussed languages. 

\section*{Acknowledgments}
We want to thank all the anonymous reviewers as well as Pavel Denisov for their helpful comments and suggestions. This project has benefited from financial support to Manuel Mager by a DAAD Doctoral Research Grant.
\bibliography{anthology,custom}
\bibliographystyle{acl_natbib}

\newpage 
\clearpage
\appendix
\section{Appendix}
\label{sec:appendix}

\subsection{Data set splitting}
\begin{table}[h!]
    \centering
    \begin{tabular}{c|r r r}
         &  Train & Dev. & Test\\\hline
        \texttt{hch}--\texttt{spa} & 665 & 167 & 553  \\
        \texttt{nah}--\texttt{spa} & 540 & 134 & 449 \\
        \texttt{tar}--\texttt{spa} & 604 & 163 & 329  \\
        \texttt{shp}--\texttt{spa} & 504 & 136 & 274 
    \end{tabular}
    \caption{Data splitting (in number of instances) used for out the Morphological Segmentation experiments for all languages.}
    \label{tab:data_splitting}
\end{table}

\begin{table}[h!]
    \centering
    \begin{tabular}{c|r r r}
         &  Train & Dev. & Test\\\hline
        \texttt{hch}--\texttt{spa} & 7442 & 447 & 1075 \\
        \texttt{nah}--\texttt{spa} & 14208 & 644 & 1291 \\
        \texttt{tar}--\texttt{spa} & 12987 & 582 & 1021 \\
        \texttt{shp}--\texttt{spa} & 13102 & 587 & 1030
    \end{tabular}
    \caption{Data splitting (in number of phrases) used for out Machine Translation experiments, from and to Spanish.}
    \label{tab:data_splitting}
\end{table}

\subsection{The Languages of new collected datasets}
\paragraph{Raramuri} (also known as Tarahumana) is a Yuto-Aztecan language, spoken in the northern part of the Mexican Sierra Madre Occidental by 89,503 speakers \cite{INGEI2020censo}. Raramuri is a polysynthetic and agglutinative language and has a Subject-Object-Verb (SOV) word order with morphonological fusion indicated by verbal suffixes \cite{caballero2008choguita}.

\paragraph{Shipibo-Konibo} is a Panoan language spoken by around 26,000 people in the Amazonian region of Perú. This language is polysynthetic, with a strong tendency to agglutination, but also with certain degree of fusion. Its word order is mainly SOV \cite{dixon1999amazonian}.

\subsection{Additional related work}

Morphological segmentation was firs introduced by \newcite{harris1951methods}. Unsupervised methods are popular with the Morfessor \cite{creutz2002, creutz2007unsupervised, poon2009unsupervised} family of segmentors. They also have a semi-supervised version \cite{kohonen2010semi, gronroos2014morfessor}. Recently Adaptor Grammars have been applied with great success to the task \cite{eskander-etal-2019-unsupervised,eskander-etal-2020-morphagram}. 
Supervised methods have achieved the best results with methods like CRFs \cite{ruokolainen2013supervised}, LSTM taggers \cite{wang2016morphological}, seq2seq RNNs \cite{kann2018fortification}, CNNs \cite{sorokin-2019-convolutional},  pointer networks \cite{yang2019point}, and pointer generator networks \cite{mager-etal-2020-tackling}. 

For the MT down-stream task, few research has been done \cite{schwartz2020neural,roest-etal-2020-machine}. New research has been done in context of the WMT 2020 shared task on Inuktitut-English \newcite{bawden2020university,kocmi2020cuni,knowles-etal-2020-nrc,roest-etal-2020-machine}. 

\subsection{Machine translation results}

Table \label{tab:mt} shows the translation results using BLEU\footnote{BLEU + case.mixed + numrefs.1 + smooth.exp + tok.13a + v.1.5.0} and chrF\footnote{chrF2 + numchars.6 + space.false + v.1.5.0}.

\begin{table*}[b]
    \centering
    \begin{tabular}{c | r r | r r | r r | r r }
        system&\multicolumn{2}{c}{hch-spa}&\multicolumn{2}{c}{nah-spa}&\multicolumn{2}{c}{tar-spa}&\multicolumn{2}{c}{shp-spa}\\\hline
                   & BLEU & chrF~~ & BLEU & chrF~~ & BLEU & chrF~~ & BLEU & chrF~~ \\\hline
bpe       & 15.04~~ & 30.50~~ &\bf 15.37~~ &\bf 37.63~~ & 11.44~ & 32.64~~ & 11.85~~ & 32.59~~ \\
morfessor & 15.12~~ & 29.23~~ &  13.84* & 35.29*  & 12.05~ & 30.35* & 9.65* & 30.69* \\
flatcat   & 15.89* &\bf 31.44* & 14.89~~ & 36.99* &\bf 15.55* &\bf 35.09* & \bf 12.29~~ & \bf 33.38~~ \\
lmvr      &\bf 16.61* & 30.96~~ & 14.78* & 36.76* & 12.97* & 33.93* & 11.14~~ & 32.60~~ \\
crfs      & 10.66* & 26.45* & 12.48* & 32.81* & 8.42* & 24.38* & -~~ & -~~ \\
seq2seq   &  9.23* & 26.64* & 12.13* & 34.62* & 7.69* & 27.10* & 10.27* & 31.10* \\
seq2seq-rand-mt & 11.46* & 26.77* & 12.22* & 32.93* & 8.31* & 24.79* & 9.51* & 27.79* \\
pointernet & 10.33* & 25.49* & 11.78* & 32.16* & 7.85* & 23.46* & 8.91* & 27.97* \\ \hline\hline
        system&\multicolumn{2}{c}{spa-hch}&\multicolumn{2}{c}{spa-nah}&\multicolumn{2}{c}{spa-tar}&\multicolumn{2}{c}{spa-shp}\\\hline
                   & BLEU & chrF~~ & BLEU & chrF~~ & BLEU & chrF~~ & BLEU & chrF~~ \\\hline
bpe     & 16.98~~ & 31.18~~ & \bf 13.29~~ & \bf 40.25~~ & 10.70~~ & 29.60~~ & 10.84~~ & 36.54~~ \\
morfessor & 12.26* & 29.60* &  8.52* & 35.55* & 5.95* & 26.72* & 5.00* & 33.24* \\
flatcat & \bf 18.70* & \bf 35.12* & 12.42* & 39.59* & 8.66* & 29.52~~ & 11.68~~ & 37.58~~ \\
lmvr    & 17.44~~ & 33.79* & 12.26* & 40.11~~ & \bf 12.88* & \bf 33.76* & \bf 12.84~~ & \bf 38.99* \\
crfs    & 9.37* & 25.40* & 6.41* & 30.33* & 2.27* & 16.28* & -~~ & -~~ \\
seq2seq & 9.64* & 28.48* & 1.29* & 24.62* & 2.96* & 24.06* & 0.77* & 25.21* \\
seq2seq-rand-mt & 7.76* & 26.11* & 3.79* & 25.54* & 1.16* & 16.29* & 0.13* & 22.06* \\
pointernet & 4.22* & 23.22* & 2.76* & 25.77* & 0.76* & 13.97* & 0.06* & 22.78* \\

    \end{tabular}
    \caption{Translation results on test for both directions. Maximum scores are in bold. We run a paired approximation test with 10000 trials using the \BPEs~ system output as the baseline, and ``*'' indicates a p-value $<$ 0.05.}
    \label{tab:mt}
\end{table*}

\end{document}